\documentclass[a4paper,fleqn]{cas-dc}

\usepackage[numbers]{natbib}

\usepackage[most]{tcolorbox}
\usepackage{caption}

\usepackage{algorithm}
\usepackage[noend]{algpseudocode}
\usepackage{amsmath}
\usepackage{amsfonts}

\algblock{Input}{EndInput}
\algnotext{EndInput}
\algblock{Output}{EndOutput}
\algnotext{EndOutput}

\begin{document}

\let\WriteBookmarks\relax
\def\floatpagepagefraction{1}
\def\textpagefraction{.001}
\shorttitle{ }
\shortauthors{D. Mikhailapov et~al.}

\title [mode = title]{
Semi-Supervised 3D Segmentation for Type-B Aortic Dissection with Slim UNETR
}

\author[1]{Denis~Mikhailapov}
\ead{d.mikhailapov@g.nsu.ru}
\author[1]{Vladimir~Berikov}[style=chinese]
\ead{berikov@math.nsc.ru}

\affiliation[1]{organization={Sobolev Institute of Mathematics SB RAS},
                city={Novosibirsk},
                country={Russia}
}

\begin{abstract}
Convolutional neural networks (CNN) for multi-class segmentation of medical images are widely used today. Especially models with multiple outputs that can separately predict segmentation classes (regions) without relying on a probabilistic formulation of the segmentation of regions. These models allow for more precise segmentation by tailoring the network's components to each class (region). They have a common encoder part of the architecture but branch out at the output layers, leading to improved accuracy.

These methods are used to diagnose type B aortic dissection (TBAD), which requires accurate segmentation of aortic structures based on the ImageTBDA dataset, which contains 100 3D computed tomography angiography (CTA) images. These images identify three key classes: true lumen (TL), false lumen (FL), and false lumen thrombus (FLT) of the aorta, which is critical for diagnosis and treatment decisions. In the dataset, 68 examples have a false lumen, while the remaining 32 do not, creating additional complexity for pathology detection.

However, implementing these CNN methods requires a large amount of high-quality labeled data. Obtaining accurate labels for the regions of interest can be an expensive and time-consuming process, particularly for 3D data. Semi-supervised learning methods allow models to be trained by using both labeled and unlabeled data, which is a promising approach for overcoming the challenge of obtaining accurate labels. However, these learning methods are not well understood for models with multiple outputs.  

This paper presents a semi-supervised learning method for models with multiple outputs. The method is based on the additional rotations and flipping, and does not assume the probabilistic nature of the model's responses. This makes it a universal approach, which is especially important for architectures that involve separate segmentation.

\end{abstract}


\maketitle

\section{Introduction}

Type B aortic dissection (TBAD) is a serious medical complication that affects the heart and blood vessels. It occurs when the inner layer of the aorta is damaged, it can lead to the formation of two separate blood channels: the true lumen (TL) and the false lumen (FL). This condition can be life-threatening if left untreated.
Early detection and treatment of TBAD are crucial for patient survival. Every year, approximatel 3–4 people per 100,000 population are affected by this disease, and about 20\% of them will die without proper medical care \cite{karthikesalingam2010diagnosis}.

There are two main treatment options for TBAD: medical therapy and thoracic endovascular intervention. If medical treatment is started early and the diagnosis is accurate, the mortality rate for patients within 30 days can be significantly reduced, to less than 10\%. \cite{hagan2000international}.
Accurately identifying the affected area is crucial for diagnosing TBAD. It helps determine the effectiveness of surgical procedures and manage the patient's condition after surgery.

Computed tomography angiography (CTA) is a commonly used imaging technique for diagnosing and treatment of type B aortic dissection. Segmentation of the true and false lumens is a crucial step in diagnosis, but manual slice-by-slice segmentation is labor-intensive and challenging. For example, some images may have a complex structure or have poor contrast. This can lead to challenging situations that require additional methods for training segmentation models. It highlights the importance of automated segmentation techniques.

Some rule-based strategies have been proposed for the segmentation of type B aortic dissection and have shown promising results with good performance \cite{fetnaci20133d, morais2017competitive}. However, these methods still require human intervention, and the segmentation results lack detailed 3D information.

With the advent of large annotated datasets, deep learning methods have demonstrated the potential for accurate segmentation of medical images \cite{yang2022weakly, dobshik2023acute}. Chen et al. \cite{chen2021multi} introduces a novel multi-stage deep learning framework for accurately segmentation of aortic dissection type B (the true lumen, false lumen, and branches). To simplify the anatomy of the curvature prior to the following segmentation, a method for straightening the aorta was developed, which based on the previous vascular anatomy in aortic dissection, allowing for the simplification of the curved shape of the aorta before creating a second network. In work \cite{kesavuori2023deep} 2D and 3D convolutional neural networks were used to segment the entire aorta. The results showed that 2D CNN surpassed 3D CNN in terms of the accuracy of highlighting the outer surface, while the indicators were equal in terms of overall segmentation accuracy.

In paper Cao et al. \cite{cao2019fully}, the authors consider the problem of segmentation of the true lumen, false lumen and the entire aorta. They explore 3 segmentation options: using  separate binary segmentation models for each classes (single one-task), branching the architecture into separate class segmentation blocks at the end of the network (single multi-task), and segmenting the entire volume to using for segmenting the main classes  (serial multi-task). The authors have demonstrated that the best results are achieved by a serial multi-task model. This model consists of two subnets, which combines the ideas of a single-task and multi-task models.

However, all these approaches require the collection and annotation of large labeled (segmented) data for network training, in contrast, it is much easier to collect a large set of unlabeled (or "weak-labeled") data and use special learning methods. Semi-supervised learning (SSL) methods make it possible to use unlabeled data using data augmentations and special additional segmentation models.

In this paper, we propose a new Semi-supervised learning method for serial multi-task modal training. The model learns to extract useful information from the data on its own, with decrease impact of inaccurate segmentation masks, as well as reduces the cost of developing a segmentation model.

In our approach, we propose special preprocesing for ImageTBAD data to improve the multi-task model's performance with unlabeled data. Our training includes spatial transformations like random rotation and flipping for data augmentation and generates pseudo-labels using an Exponential Moving Average (EMA) model.

It is worth to note that all the works that use TBDA segmentation methods use private datasets and not multi-output models. While we are using an open dataset with 100 labeled examples. For this reason, we do not aim to compare our results with other works, but rather focus on the development and research of a semi-supervised segmentation method.

The rest of this manuscript is organized as follows: In the next section, we observe the related work, specifically pseudo-labeling and consistency regulation methods. Section \ref{mat_and_met} describes the materials and methods. In particular, \ref{dataset} and \ref{preproc} describe the dataset and the preprocessing method, \ref{model} and \ref{metric} describe the model and the evaluation metric, \ref{method} describe our semi-supervised learning method, and \ref{Result} describe experiments. Results and conclusion in section \ref{Result} and 5, respectively.

\section{Related work}
\label{rel_work}

Semi-supervised learning (SSL) is an important area of research for medical image segmentation, where labeled data is often expensive. Various SSL methods have been developed, which can be categorized into five main approaches: pseudo-labeling, consistency regularization, contrastive learning, GAN-based methods, and hybrid approaches \cite{han2024deep}. Each method offers unique mechanisms and advantages that contribute to enhancing model performance in segmentation tasks.

In this section, we mainly describe pseudo-labeling and consistency regularization approaches, because our method based on it.

\subsection{Pseudo-Labeling}

A fundamental approach in SSL is pseudo-labeling, which involves using an initial model (or its modification) trained on the basis of labeled data to create pseudo-labels for unlabeled samples. This allows iteratively expanding the labeled data set using predictions with high accuracy. In this context, such methods as self-learning and co-training are widespread.

The process often incorporates self-training \cite{wang2021few, chaitanya2023local, qiu2023self, wu2023compete, xu2024expectation}, where a model iteratively refines its predictions by retraining on the pseudo labels it generates. A critical aspect of this method is the quality of the pseudo labels; poor initial predictions can introduce noise and negatively impact the model's learning.

Alternatively, co-training \cite{xia20203d, peng2020deep, xia20203d} is employed, where multiple models are trained simultaneously on different but complementary views of the data. Each model shares high-confidence predictions with the others, fostering a collaborative learning environment that can improve robustness but increases computational costs.

Improving the quality of pseudo labels is essential. Techniques such as combining outputs from different models or using uncertainty estimation can help ensure that the pseudo labels are reliable. However, the effectiveness of pseudo-labeling is significantly dependent on the initial model's accuracy; if noisy predictions are generated, subsequent training can suffer.

Also in this context, Exponential Moving Average (EMA) technique is employed \cite{cai2021exponential}. It helps create more stable and reliable pseudo-labels by aggregating the model during the training process. Instead of using only the most recent model predictions, EMA creates an aggregated model that reflects both current and previous states. This approach reduces noise and improves the quality of pseudo-labels in semi-supervised learning scenarios.

\subsection{Consistency Regularization}

Another semi-supervised learning method is consistency regularization \cite{qiao2022semi, wang2023mcf, gao2023correlation}, which asserts that a model's predictions should remain stable despite variations or perturbations in the input data. This concept rests on the smoothness hypothesis, which posits that small modifications to the input should not result in substantial changes in the model's output. Consistency regularization can effectively impose prior constraints that help mitigate the risks associated with relying on pseudo-labels, which may lead to training instability.

In the paper by Oliver et al. \cite{oliver2018realistic}, the authors explore the concept of consistency regularization, highlighting its application in three key areas. 
Firstly, data consistency focuses on maintaining stable predictions when the unlabeled data is transformed or randomly perturbed. The authors argue that the model's output should remain unchanged despite variations in the input.
Secondly, model consistency emphasizes the need for the same input to produce consistent output across different models. This led to the development of approaches like the Mean Teacher model \cite{tarvainen2017mean}, which aims to achieve this consistency.
Lastly, task consistency stresses the importance of producing uniform outputs across different tasks. This allows the model to utilize a broader range of information, enhancing its overall robustness and performance.
By focusing on these three aspects, the process of regularizing consistency offers a holistic approach to improving the dependability and steadiness of model outcomes in the face of input variation.

\subsection{Other approaches}

In addition, there are other SSL methods to teaching medical segmentation. Contrastive learning \cite{hu2021semi, pandey2021contrastive, wang2022semi} aims to maximize the similarity between related instances and minimize it between dissimilar ones, which allows models to recognize anatomical structures. Generative adversarial networks (GANs) \cite{tan2022semi, wang2022generative} create realistic samples by training the discriminator to distinguish between real and generated data, which is especially useful when labeled data is insufficient. Hybrid methods combine the strengths of these methods by applying consistency regularization and uncertainty estimation, increasing the reliability of the model. The diversity of these strategies highlights the importance of research in this area, opening up new opportunities for analyzing medical images and improving clinical outcomes. 

\section{Material and Methods}
\label{mat_and_met}

In this section, we describe the materials and methods employed for the supervised learning system focused on 3D segmentation. The model is trained on both labeled and unlabeled data, addressing challenges in obtaining accurate labeled data in medical imaging. To increase dataset diversity, spatial transformations, such as random rotation and flipping, were applied, while pseudo-labels were generated using the EMA model.

Data preprocessing involved resizing images to a voxel size of 128 x 128 x 128 and applying voxel intensity transformations to accentuate important structures for enhanced segmentation.

Model evaluation utilized the DICE coefficient to measure agreement between predicted segmentation masks and expert labels. We conducted five experiments with varying proportions of labeled data, assessing the impact of incorporating unlabeled data with and without augmentation. This approach aimed to improve the accuracy and effectiveness of 3D segmentation techniques for imaging type B aortic dissection.

\subsection{Dataset}
\label{dataset}

The ImageTBAD dataset \cite{yao2021imagetbad} consists of 100 three-dimensional computed tomography angiography (CTA) images obtained between January 1st, 2013 and April 23rd, 2015. All these images are preoperative images of thoracic brachiocephalic artery disease (TBAD), with the upper and lower portions located around the neck and brachycephalic vessels, respectively, in an axial view. Segmentation labeling was performed by a team of experienced cardiologists and radiologists who have extensive experience working with TBAD cases. The segmentation includes three substructures: true lumen (TL), false lumen (FL), and false lumen thrombus (FLT). There are 68 images that contain FLT, and 32 that do not. Previous studies have found that the FLT class is difficult to segment. The best accuracy achieved was approximately 0.52, as measured by the DICE score.

The main purpose of our article is to study the semi-supervised learning method, so we will simplify the task of data segmentation in order to better explore our method. To train the model, a new class was formed, which is the union of three substructures. Denote it conditionally as class "ALL". Three segmentation classes were used in the final set: ALL, TL, FL.

\subsection{Data preprocessing}
\label{preproc}

The source data was reduced in size to 192 pixels in the XY plane. Then, 32 pixels were removed from the image border in the same plane. As a result, the final image size was 128 by 128 pixels along the X and Y axes, and the Z axis was also reduced to 128 pixels. The final size of each 3D image became equal to (128, 128, 128).

\begin{figure}[!ht]
    \includegraphics[width=0.88\linewidth]{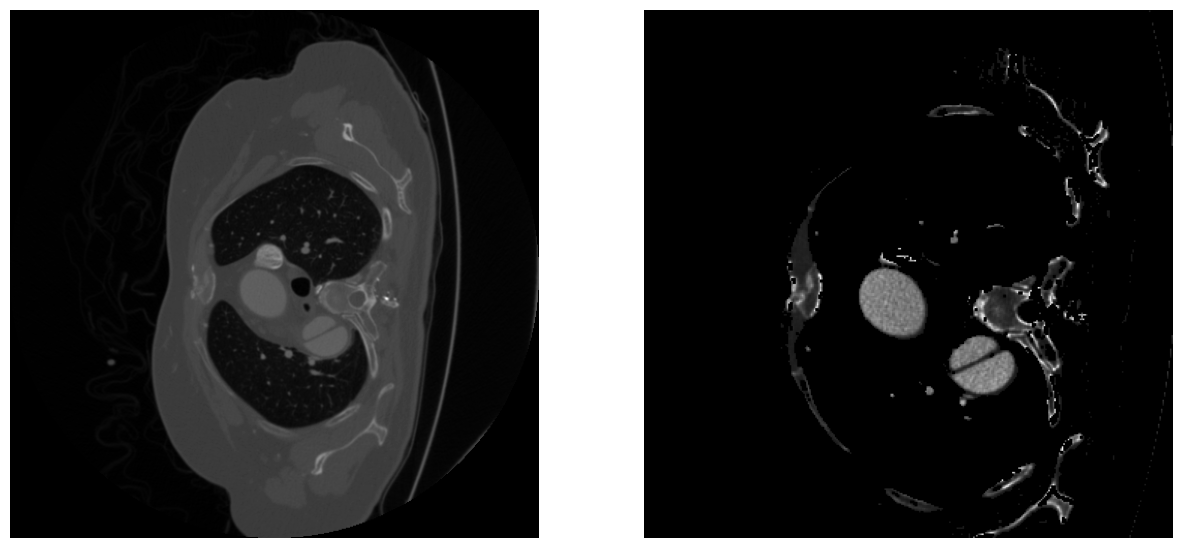}
    \centering
    \caption{An example of data preprocessing, before and after.}
    \label{fig:data_preproc_example}
\end{figure}

There was also an intensity restriction on the Hounsfield scale, the application of standardization, exponential functions and final MinMax normalization. This helps to highlight important structures in the image and facilitate the segmentation process. An example of image processing code is shown in Listing~\cite{lst:preprocess}. An example of the resulting image can be seen in Fig~\ref{fig:data_preproc_example}.

\begin{lstlisting}[
  language=Python,
  caption={Image preprocessing code},
  label={lst:preprocess},
  basicstyle=\ttfamily\small,
  keywordstyle=\color{blue},
  commentstyle=\color{green},
  stringstyle=\color{red}
]
import numpy as np
from scipy import ndimage

def process_image(
    data_img: np.ndarray, 
    vmin=1100, 
    vmax=1600, 
    exp_coef=1.3
):
    data_img[data_img < vmin] = 0
    data_img[data_img > vmax] = 0
    mask = ndimage.grey_erosion(data_img, size=(2, 2, 1))
    data_img[mask == 0] = 0

    data_img -= 2 * np.mean(data_img)
    data_img /= np.std(data_img + 1e-6)
    data_img = np.exp(exp_coef * data_img)
    data_img = data_img - data_img.min()
    data_img /= (data_img.max() - data_img.min())
    
    return data_img
\end{lstlisting}

\subsection{Model}
\label{model}

\begin{figure}[!ht]
    \includegraphics[width=1.0\linewidth]{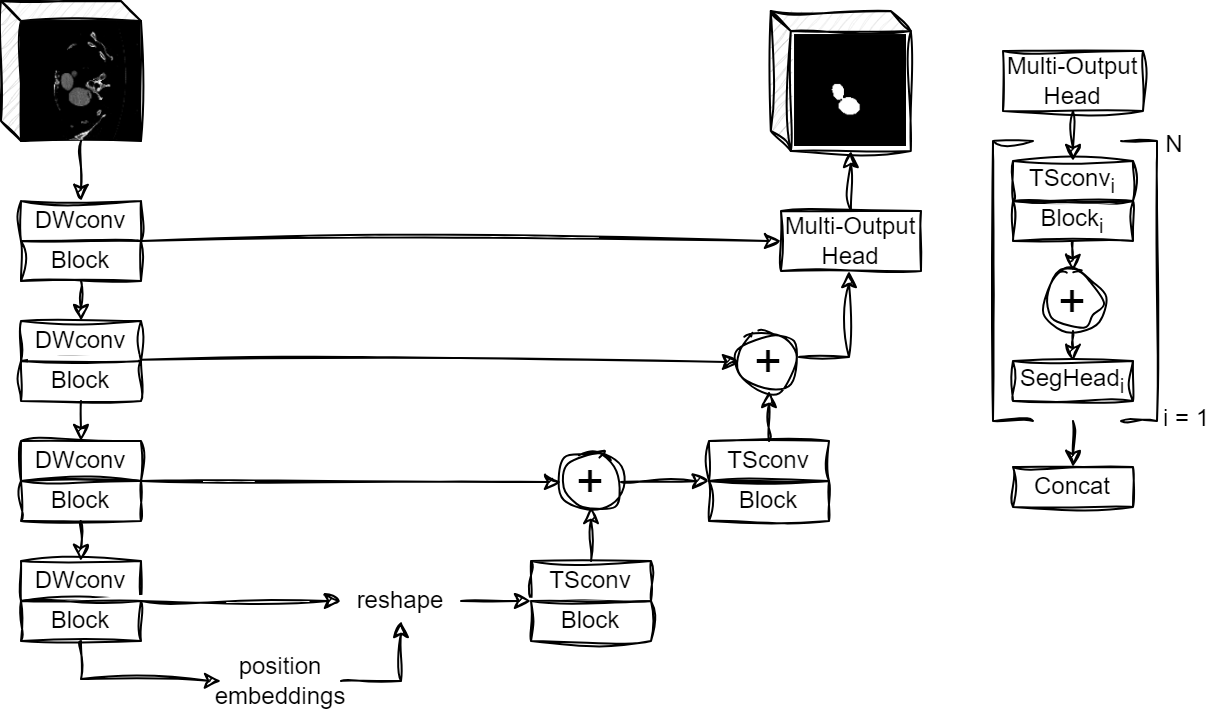}
    \centering
    \caption{Multi-Output Slim UNETR Architecture.}
    \label{fig:Model_Slim_unetr_Multi_Output}
\end{figure}

In this work, we used Slim UNETR model \cite{pang2023slim} with multi-output head (Fig. \ref{fig:Model_Slim_unetr_Multi_Output}). The architecture consists of three main components: the down-scale part, the bottleneck and the up-scale part. The down-scale part includes several convolutional blocks that are used in the Slim UNETR architecture, then bottleneck applies positional encoding to downscale feature maps, and finally, the up-scale part decodes the segmentation mask from the feature maps. A key feature of the multi-output architecture is the use of separate output blocks for each class in a segmentation mask. This can improve segmentation accuracy because it blocks the specialized segmentation mask of the decoding class, but the resulting segmentation masks are not adjusted relative to each other.

\subsection{Metric of Model quality}
\label{metric}

The quality of predictions made by the constructed models was evaluated using the DICE coefficient:
\begin{equation}\notag
DICE_c\left(y,\hat{y}\right)=\frac{2 \sum_{i,j,k}\hat{y}_{i,j,k,c}\cdot y_{i,j,k,c}+\varepsilon}{\sum_{i,j,k}\hat{y}_{i,j,k,c}^{2}+\sum_{i,j,k} y_{i,j,k,c}^{2}+\varepsilon}
\end{equation}

\noindent where $y=\left\{ y_{i,j,k,c}\right\}$ corresponds to the segmentation mask delineated by an expert;
$\hat{y}=\left\{ \hat{y}_{i,j,k,c}\right\}$ is the predicted mask generated by the model; $i,j,k$ are voxel indices, $c$ is class index and $\varepsilon$ is a small positive value.

\subsection{Proposed Method}
\label{method}

In this section, we will introduce our proposed semi-supervised learning method. Self-supervised learning (SSL) is a machine learning paradigm that combines supervised and unsupervised learning elements. 

In supervised learning, labeled 3D samples are utilized to train the model for accurate prediction of 3D segmentation regions. To optimize the training process, we employ Generalized Dice Loss (GDL) and Focal Loss (FL).

During unsupervised the training process, pseudo-labels are generated using an EMA model and data augmentations. This EMA model uses the weight parameters of the main model.

To summarize, the semi-supervised machine learning task can be written as follows:
\begin{equation} \notag
    \begin{array}{l}
        D_l = \{(x_l, y_l)\} - \textit{labled data}  \\
        D_u = \{(x_u)\} - \textit{unlabled data}  \\
        \\
        f_\theta(x_{l, u}) = \hat{y}_{l, u} - \textit{model prediction}
        \\
        f_{\theta^{ema}}(x_u) = y_p - \textit{model pseudo-labels prediction}
        \\
        \theta^{ema}_t = \mu \theta^{ema}_{t-1} + (1 - \mu) \theta_t
        \\
        \\
        L_{total} = L_{label}(f_\theta,\, D_l) + L_{unlab}(f_\theta,\, D_u)
    \end{array}
\end{equation}

The details of the implementation of the method are presented in sections \ref{sup_part} and \ref{unsup_part}.

\subsubsection{Supervised learning} 
\label{sup_part}

The weights of the neural network are optimized by minimizing a loss function using the backpropagation algorithm. However, the data in the dataset may be unbalanced, which means that some classes are underrepresented. For example, The FLT class is only present in 68 out of the 100 examples in our dataset, and the TL and FL classes also have different numbers of examples. It is therefore important to use loss functions that are able to account for this class imbalance.

In our work, we use two loss functions:
GeneralizedDiceLoss (GDL)~\cite{sudre2017generalised} and FocalLoss (FL)~\cite{ross2017focal}. The study~\cite{chen2017deeplab} shows that using a mixed loss function can improve the quality of the final model.

To directly enhance the Dice metric, we use the Generalized Dice loss function:
\begin{equation} \notag
    L_{GDL}(y,\hat{y})=1 - 2\frac{\sum_{c}w_c \sum_{i,j,k}\hat{y}_{i,j,k, c}\cdot y_{i,j,k, c} + \varepsilon}{\sum_{c}w_c (\sum_{i,j,k, c}\hat{y}_{i,j,k, c}^{2}+\sum y_{i,j,k, c}^{2}) + \varepsilon}.
\end{equation}

where $i,j,k$ are the voxel indices, $c$ is the class index, $w_c = 1/(\sum_{i,j,k} y_{i,j,k, c})^{2}$.

To work with unbalanced classes, we use Focal loss function: 
\begin{equation} \notag
   \begin{split}
        L_{FL}(y,\hat{y})=-\sum_{i,j,k, c} \alpha_c [y_{i,j,k, c}(1-\hat{y}_{i,j,k, c})^{\gamma} \cdot \\ \cdot log(\hat{y}_{i,j,k, c})+ (1-y_{i,j,k, c})\hat{y}_{i,j,k, c}^{\gamma}\cdot log(1-\hat{y}_{i,j,k, c})]
    \end{split}
\end{equation}

The hyperparameter $\gamma$ addresses the issue of class imbalance in the dataset. A higher value of $\gamma$ reduces the impact of easy-to-classify examples on the training process, while a lower value increases it. In addition, we use a class weight multiplier $\alpha_c$ to improve accuracy for small classes.

As the final supervised loss function, a weighted average combination of $L_{FL}$ and $L_{GDL}$ was used:
\begin{equation} \notag
    L(y,\hat{y})=w_{GDL} L_{DICE}(y, \hat{y}) + w_{FL} L_{FL}(y, \hat{y}).
\end{equation}
\noindent 

\subsubsection{Unsupervised learning} 
\label{unsup_part}

The process of unsupervised learning is similar to supervised learning, but it does not require ground truth samples. 

\textbf{EMA model:}
To obtain these ground truth (pseudo-labels), the EMA model is used: First, create a copy of the main model, including its architecture and parameters. Then, during each training epoch, the EMA model predicts the labels for the unlabeled samples, producing the pseudo-labels and then updates its own parameters using the main model's parameters according to the following formula: $$\theta^{ema}_t = \mu \theta^{ema}_{t-1} + (1 - \mu) \theta_t$$ where $\theta^{ema}, \theta$ parameters of models, \( \mu \) represents a smoothing factor that usually approximately equal 0.95.

This formula can be rewritten as follows: $$\theta^{ema}_t = \theta_t +  \mu (\theta^{ema}_{t-1} - \theta_t)$$
here we can note that the term $\theta^{ema}_{t-1} - \theta_t$ is the delta of the weights of the models. Hence the parameter \( \mu \) adjusts the delay rate of the model.

The EMA approach provides several advantages for predicting pseudo-labels in semi-supervised learning. It stabilizes the training process, reducing noise and enhancing model generalization. Additionally, the EMA model provides consistent and reliable targets for pseudo-labeling by acting as a form of models aggregation. Its smoothing effect also helps reduce overfitting, which is particularly beneficial when working with limited labeled data.

\textbf{The full algorithm for generating pseudo-labels.:}
The general algorithm for generating pseudo-labels can be described as follows:

\begin{enumerate}
    \item First, applying random rotation and flip augmentations to the unlabeled image. Then, the prediction of the model is calculated ($\hat{y}_u$).
    \item After that, the pseudo-label is calculated using the EMA model. And then, the same augmentations are applied to this pseudo-label without randomness. Finally, it is binarized, so that each pixel has a value of 0 or 1 ($y_p$).
    \item At the end, the loss function is calculated as in the supervised part.
\end{enumerate}

\begin{figure}[!ht]
    \includegraphics[width=1.0\linewidth]{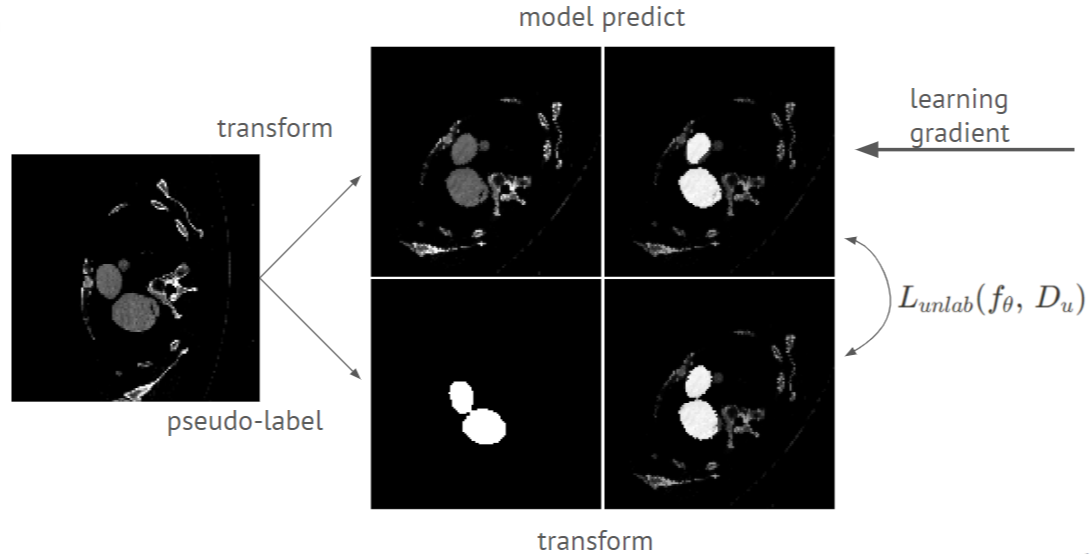}
    \centering
    \caption{The scheme of the semi-supervised method.}
    \label{fig:method_overview}
\end{figure}

The Figure \ref{fig:method_overview} shows a diagram of the method.

Note that unlabeled data is used in the learning process not from the beginning of learning, but after a certain epoch. This is necessary for the model to learn the initial understanding of the data and the segmentation tasks. 

The summary of our semi-supervised learning method is presented in Algorithm \ref{alg:semi}.

\begin{algorithm}[h!]
  \scriptsize
\caption{\small{Semi-Supervised Learning}}

\textbf{Input}: \\
  $D_L = \{(x^L_b, y^L_b) \mid b \in \{1, \dots, N\}\}$ — labeled dataset \\
  $D_U = \{x^U_b \mid b \in \{1, \dots, M\}\}$ — unlabeled dataset \\
  $T_s$ — Spatial random transform with random seed $s$ \\
  $T$ — Number of epochs \\
  $T_{\text{unlab\_start}}$ — Start unlabeled train \\
\textbf{Output}: $\text{Model parameters}\; \theta$
\begin{algorithmic}[l]
  \State $\theta_0$ - \text{initual parameters}
  \State $\theta^{ema}_0 := \theta_0$
  \State \ForAll{$t = 1, \dots, T$}:
      \State $s^{'} := rand\_seed()$ 
      \State $L_L(\theta_{t-1},\,D_L) = \sum_{(x_l,\,y_l) \in D_L} L_{label}(f(\theta_{t-1},\,\mathcal{T}_{s^{'}}(x_l)), y_l)$
      \State $\theta_{t} := \theta_{t-1} + \eta \nabla_{\theta}L_{L}(\theta_{t-1},\,D_L)$
      \State \If{$t > \hat{T}$}: 
         \State $\theta^{ema}_{t} := no\_grad_{\theta^{ema}}(\theta_t + \mu (\theta^{ema}_{t-1} - \theta_{t-1}))$ 
         \State $s^{''} := rand\_seed()$
         \State $L_U(\theta_t,\, \theta^{ema}_t,\, D_U)  = \sum_{x_u \in D_U} L_{unlabel}(f(\theta_{t}, \mathcal{T}_{s^{''}}(x_u)), \mathcal{T}_{s^{''}}(f(\theta^{ema}_t, x_u))))$ 
         \State $\theta_{t} := \theta_{t} + \eta \nabla_{\theta}L_U(\theta_{t}, \theta^{ema}_t, D_U)$
      \State \EndIf
  \State \EndFor
  \State \textbf{Return} $\theta_{T}$
\end{algorithmic}
 \label{alg:semi}
\end{algorithm}

\subsection{Experiments performed}
\label{exp_perf}

In this study, the following experiments were performed:
\begin{enumerate}
  \item A full labeled dataset
  \item Training on $50\%$ of the labeled data
  \item The method of semi-supervised learning based on $50\%$ of the labeled data
  \item A method of semi-supervised learning based on $50\%$ labeled data with augmentation of labeled data
\end{enumerate}

\textbf{Implementation details:} The number of training epochs is 800, the size of the batch is 6, $\gamma = 2., w_{GDL} = 1., w_{FL} = 0.8$, a class weight multipliers $\alpha = [0.8, 0.9, 1.5]$, for each class, respectively. Data splitting into training and validation is 80/20\%, AdamW is used as an optimizer. The training code is originally taken from the official implementation of Slim-UNETR on github\footnote{https://github.com/aigzhusmart/Slim-UNETR}. The training was conducted on an Nvidia Tesla P100 16gb GPU.

\section{Results}
\label{Result}

Table \ref{tabel:metrics} presents the DICE metrics for three distinct classes. We validated our model through cross-validation using the Shuffle\&Split method. This approach involves randomly shuffling the dataset before partitioning it into training and validation subsets. In our study, we performed this shuffling and splitting process four times. The “$\pm$” symbol represents the standard deviation of the DICE metric across these four trials. The first column of the table indicates the experiment number, while the subsequent columns display the corresponding metrics for each class.

As the results showed, the supervised learning method shows approximately the same results as training on all data. We conducted a comparison of training on all data and only on 50\% data. It is clearly shown that the lack of data affects the accuracy of the model. The behavior of the metrics is approximately the same until the 100th epoch, then the discrepancy begins \cite{shwartz2017opening}. This means that around this epoch, the drift phase of neural network optimization ends and the diffusion phase begins. And, as expected, the use of additional unlabeled data provides an increase in the DICE metric, especially on ’false lumen’ (FL) class. 

\begin{figure}[!ht]
    \includegraphics[width=1.0\linewidth]{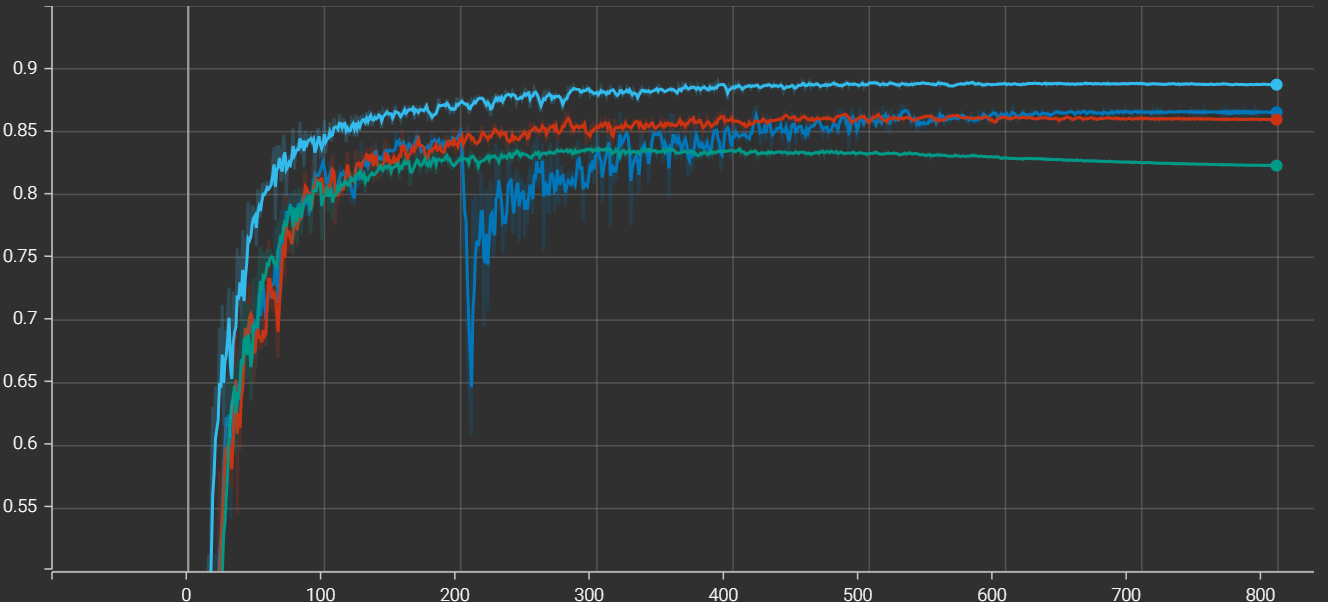}
    \centering
    \caption{An example of the growth of the DICE (ALL) metric in the learning process (same random seed): red – experiment 1, green – experiment 2, blue – experiment 3, light blue – experiment 4}
    \label{fig:training_process}
\end{figure}

It is worth noting an interesting fact: when adding unlabeled data, the metrics drop. This is probably due to the fact that when unlabeled examples are added to the learning process, the model is forced to learn from new data. Therefore, the model requires more time for full training. This can be seen in Figure \ref{fig:training_process}, at the 200th epoch. However, at the end of training, the metric rises to the level of the full data set.

\begin{table}[!ht]
    \centering
    \caption{Class-wise DICE score for different experiment settings}
    \label{tabel:metrics}
    \begin{tabular}{| c | c | c | c |} \hline
        № & ALL              &  TL              & FL               \\ \hline
        1 & 86.51\tiny{±0.28} & 76.32\tiny{±0.54} & 60.84\tiny{±0.47} \\ \hline
        2 & 83.04\tiny{±0.3} & 72.2 \tiny{±0.48} & 58.04\tiny{±0.36} \\ \hline
        3 & 85.82\tiny{±0.51} & 77.36\tiny{±0.8} & 63.9\tiny{±0.9} \\ \hline
        4 & 87.38\tiny{±0.21} & 78.42\tiny{±0.73} & 63.37\tiny{±1.2} \\ \hline
    \end{tabular}
\end{table}

\begin{figure}[!ht]
    \includegraphics[width=1.0\linewidth]{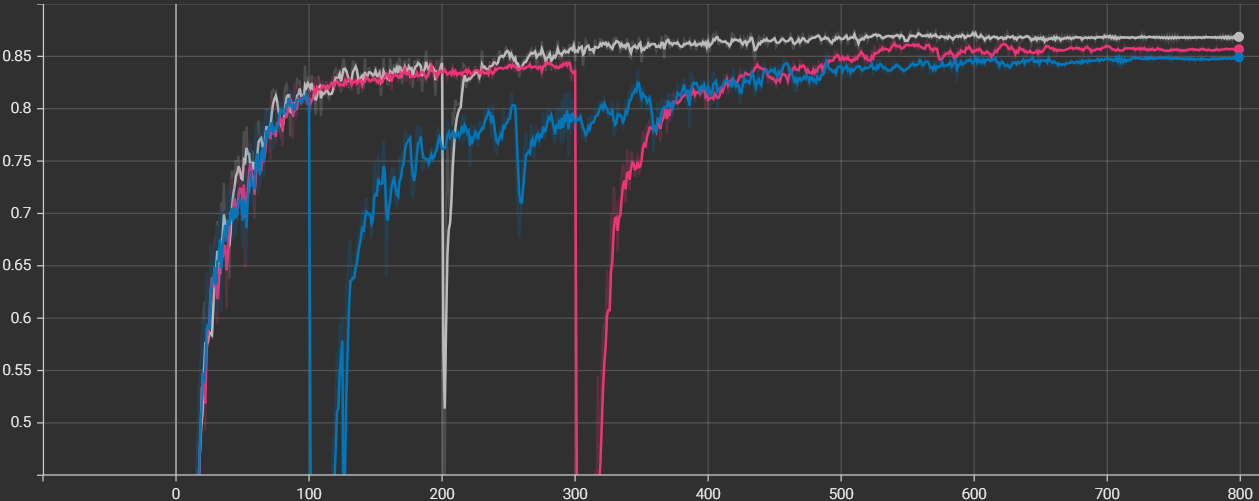}
    \centering
    \caption{An example of the growth of the DICE (ALL) metric for different start epoch for unlabeled training: blue – start epoch 100, pink – start epoch 200, gray – start epoch 300}
    \label{fig:start100_200_300}
\end{figure}

In this work, we also compared the model's accuracy for different start epochs of the semi-supervised method (see Figure \ref{fig:start100_200_300}). It turned out that, on average, the best option is to start the learning process unsupervised with epoch 200. At the start epoch of 100, the model does not have time to study the labeled data, however, at the start epoch of 300, the model no longer has time to study the unlabeled data. 

Moreover, high standard deviation of metrics makes it difficult to provide a clear answer.

\begin{table}[!ht]
    \centering
    \caption{Compare DICE score for different unlabel start epoch (100, 200 and 300)}
    \label{tabel:metrics_100_200_300_se}
    \begin{tabular}{| c | c | c | c |} \hline
        Start epoch & ALL              &  TL              & FL               \\ \hline
        100 & 85.02\tiny{±0.8} & 76.35\tiny{±0.4} & 60.92\tiny{±0.8} \\ \hline
        200 & 85.82\tiny{±0.3} & 77.36\tiny{±0.8} & 63.9\tiny{±0.8} \\ \hline
        300 & 86.15\tiny{±0.3} & 77.06\tiny{±0.6} & 61.18\tiny{±0.6} \\ \hline
    \end{tabular}
\end{table}

\section{Conclusion}
\label{Conclusion}

In this study, we introduced a novel semi-supervised learning approach for the 3D multi output segmentation model. The critical nature of accurate segmentation for TBAD diagnosis and treatment underscores the importance of our work. By leveraging both labeled and unlabeled data, our method addresses the challenges associated with obtaining extensive high-quality labeled datasets, which can be both costly and time-consuming, particularly in 3D imaging contexts.

Our approach employs data augmentation techniques, such as random rotation and reflection, alongside an Exponential Moving Average (EMA) model for generating pseudo-labels. This combination aims to mitigate the impact of inaccurate segmentation masks, enhancing the robustness of the model training. The evaluation of our method on the ImageTBAD dataset, comprising 100 CTA images, illustrates its effectiveness in improving segmentation accuracy.

Overall, our findings highlight the potential of employing semi-supervised learning strategies in medical image segmentation, demonstrating that such methods can significantly reduce the dependency on annotated data while maintaining reliable and precise segmentation outputs. In the future, it is worth testing the method on more popular datasets, as well as considering cases with a smaller percentage of labeled data.

\section*{Acknowledgment}

Author Denis Mikhailapov was supported by Sobolev Institute of Mathematics SB RAS (project FWNF-2024-0002\footnote{https://ai-biolab.ru/}) to develop and conduct experiments.
Author Vladimir Berikov received support from Russian Science Foundation (grant 24-21-00195) for experiments develop and manuscript preparation.

\printcredits

\bibliographystyle{cas-model2-names}

\bibliography{cas-refs}





\end{document}